\pdfoutput=1
\documentclass{egpubl_vcbm_short}
\usepackage{vcbm2017s}

 \WsPaper         
 \electronicVersion 

\ifpdf \usepackage[pdftex]{graphicx} \pdfcompresslevel=9
\else \usepackage[dvips]{graphicx} \fi

\PrintedOrElectronic

\usepackage{t1enc,dfadobe}

\usepackage{egweblnk}
\usepackage{cite}

\pdfoutput=1
\usepackage{amssymb}
\usepackage{tikz}

\title[{UI-Net}: Interactive Artificial Neural Networks]%
      {{UI-Net}: Interactive Artificial Neural Networks for Iterative Image Segmentation Based on a User Model}

\author[Mario Amrehn et al.]
{\parbox{\textwidth}{\centering Mario Amrehn$^{1}$, %
        Sven Gaube$^{1}$, %
        Mathias Unberath$^{1}$, %
        Frank Schebesch$^{1}$,\\
        Tim Horz$^{3}$, %
        Maddalena Strumia$^{3}$, %
        Stefan Steidl$^{1}$, %
        Markus Kowarschik$^{3}$, %
        Andreas Maier$^{1,2}$%
        }%
        \\
{\parbox{\textwidth}{\centering $^1$Pattern Recognition Lab., Computer Science Department, Friedrich-Alexander University Erlangen-Nuremberg, Germany\\
         $^2$Graduate School in Advanced Optical Technologies (SAOT), Erlangen, Germany\\
         $^3$Siemens Healthcare GmbH, Forchheim, Germany%
       }%
}
}

\begin{document}


\maketitle
\begin{abstract}
   %
   %
   For complex segmentation tasks, fully automatic systems are inherently limited in their achievable accuracy for extracting relevant objects. 
   Especially in cases where only few data sets need to be processed for a highly accurate result, semi-automatic segmentation techniques exhibit a clear benefit for the user.
   One area of application is medical image processing during an intervention for a single patient.
   We propose a learning-based cooperative segmentation approach which includes the computing entity as well as the user into the task.
   %
   Our system builds upon a state-of-the-art fully convolutional artificial neural network (\mbox{FCN}) as well as an active user model for training.
   During the segmentation process, a user of the trained system can iteratively add additional hints in form of pictorial scribbles as seed points into the FCN system to achieve an interactive and precise segmentation result.
   The segmentation quality of interactive \mbox{FCNs} is evaluated.
   Iterative \mbox{FCN} approaches can yield superior results compared to networks without the user input channel component, due to a consistent improvement in segmentation quality after each interaction.
%


\end{abstract}  

\section{Introduction}
Trans-catheter arterial chemoembolization (\mbox{TACE}) \cite{lewandowski2011transcatheter} is a minimally invasive treatment for liver cancer, utilizing image guidance.
Vessels which supply the hepatocellular carcinoma (\mbox{HCC}) with oxygenated blood are induced with chemotherapeutic 
agent and subsequently occluded by a physician \cite{kim2005recognizing}. 
During the intervention, several 2-D projection images are acquired by a cone-beam C-arm CT scanner in order to reconstruct a volumetric DynaCT image of the patient's abdomen.
A segmentation of the cancerous tissue is performed on the image data.
Subsequently, the outline of the segmented tumor is used to find collateral vessels.

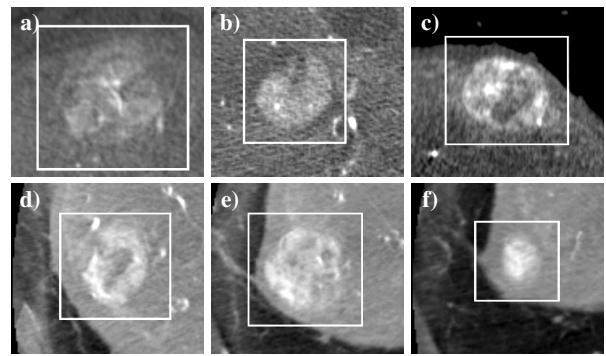
\begin{figure}
	\centering
	\resizebox{0.9829\columnwidth}{!}{%
		\Large
		{\def\arraystretch{0.75}\tabcolsep=2pt
		\begin{tabular}{lll}
			\begin{tikzpicture}
			\node[anchor=south west,inner sep=0,outer sep=0] (image) at (0,0) {%
				\includegraphics[trim={0 13 14.7244897959 0},clip,height=0.1722\textwidth,width=0.19504\textwidth]{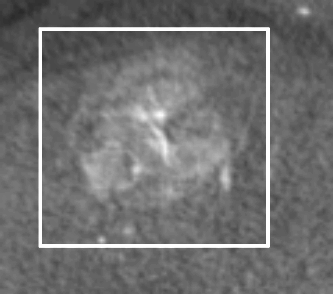}%
			};
			\begin{scope}[x={(image.south east)},y={(image.north west)}]
			\node[white,font=\bfseries] at (0.1,0.9) {a)}; 
			\end{scope}
			\end{tikzpicture} &
			\begin{tikzpicture}
			\node[anchor=south west,inner sep=0,outer sep=0] (image) at (0,0) {%
				\includegraphics[height=0.1722\textwidth,width=0.19504\textwidth]{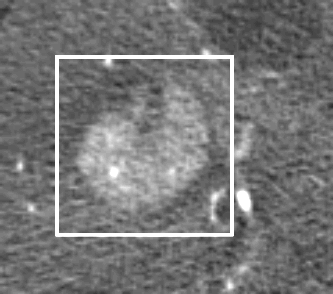}%
			};
			\begin{scope}[x={(image.south east)},y={(image.north west)}]
			\node[white,font=\bfseries] at (0.1,0.9) {b)};
			\end{scope}
			\end{tikzpicture} &
			\begin{tikzpicture}
			\node[anchor=south west,inner sep=0,outer sep=0] (image) at (0,0) {%
				\includegraphics[height=0.1722\textwidth,width=0.19504\textwidth]{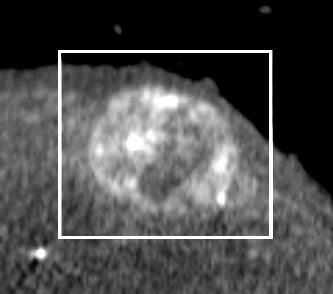}%
			};
			\begin{scope}[x={(image.south east)},y={(image.north west)}]
			\node[white,font=\bfseries] at (0.1,0.9) {c)};
			\end{scope}
			\end{tikzpicture}\\
			\begin{tikzpicture}
			\node[anchor=south west,inner sep=0,outer sep=0] (image) at (0,0) {%
				\includegraphics[trim={0 12 13.3416149068 0},clip,height=0.1722\textwidth,width=0.19504\textwidth]{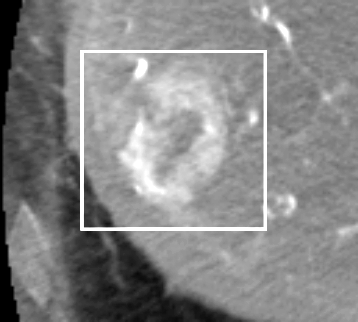}%
			};
			\begin{scope}[x={(image.south east)},y={(image.north west)}]
			\node[white,font=\bfseries] at (0.1,0.9) {d)};
			\end{scope}
			\end{tikzpicture} &
			\begin{tikzpicture}
			\node[anchor=south west,inner sep=0,outer sep=0] (image) at (0,0) {%
				\includegraphics[trim={0 12 13.3416149068 0},clip,height=0.1722\textwidth,width=0.19504\textwidth]{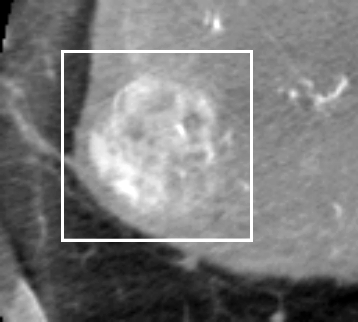}%
			};
			\begin{scope}[x={(image.south east)},y={(image.north west)}]
			\node[white,font=\bfseries] at (0.1,0.9) {e)};
			\end{scope}
			\end{tikzpicture} &
			\begin{tikzpicture}
			\node[anchor=south west,inner sep=0,outer sep=0] (image) at (0,0) {%
				\includegraphics[trim={0 12 13.3416149068 0},clip,height=0.1722\textwidth,width=0.19504\textwidth]{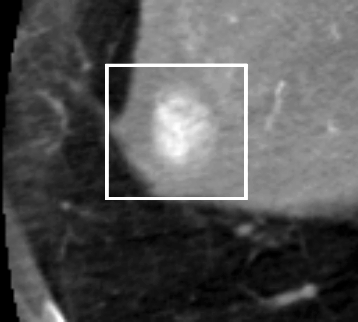}%
			};
			\begin{scope}[x={(image.south east)},y={(image.north west)}]
			\node[white,font=\bfseries] at (0.1,0.9) {f)};
			\end{scope}
			\end{tikzpicture}%
		\end{tabular}%
		}
	}%
	\caption{Challenges of hepatic lesion segmentation are (a) high diversity in gray-values, no typical shape, (b) intensity overlaps between tumor and surrounding tissue, %
		(c) intensity patches due to necrotic regions, and %
		(d--f) varying appearance of the same tumor between 2-D slices.}%
	\label{fig:hepatic_tumors_challenges}%
\end{figure}

The more accurate the segmentation, the less healthy tissue surrounding the lesion is occluded, minimizing the toxicity of the procedure.
In addition, a higher percentage of the tumor can be treated with chemotherapeutic agent, 
increasing the efficacy of the therapy \cite{lo2002randomized}.
Segmenting the tumor with high accuracy is especially challenging due to variations in shape, size (\mbox{Fig.\,\ref{fig:hepatic_tumors_challenges}} a--f), and a high diversity in X-ray attenuation (\mbox{Fig.\,\ref{fig:hepatic_tumors_challenges}} a), 
as depicted in \mbox{Fig.\,\ref{fig:hepatic_tumors_challenges}}. %

Current approaches for intra-procedural tumor segmentation can be divided into three main categories depending on 
the involvement of the operating user in the segmentation process: manual, automatic, and interactive. 
(1) With manual segmentation schemes, users draw the complete contour line of the object to be segmented with minimal assistance by the system.
A perfect manual segmentation of hepatic lesions is feasible, but would take several minutes until an appropriate result is reached during the intervention due to the primitive tools provided to the user.
(2) Fully automated segmentation approaches may also exhibit long runtimes attributable to a lack of domain knowledge of the system.
If learning-based, such methods may also need a large amount of training data in order to achieve an acceptable accuracy of their outcome.
Still, a perfect segmentation may not be reachable.
Users do not have control over the process.
However, trained physicians could substantially assist in reaching the goal of a fast and exact segmentation, due to their knowledge of a very good estimate of the true tumor extent. 
(3) Interactive segmentation methods are applicable, particularly in situations where only few or even no annotated data sets of similar segmentation tasks are available, or the task is to produce only a few new, but accurate segmentations.
The limiting factor for scaling this approach is the time spent by users to provide input during each image segmentation task.
Therefore, interactive methods are not a replacement for fully automated approaches, but can supersede them in certain niches on account of their high accuracy reached by efficient use of their operators' expertise.
\subsection{User Input}
According to \cite{olabarriaga2001interaction}, user interactions can be categorized depending on the interactive segmentation system's interface:
(1) A menu-driven user input scheme as in \cite{rupprecht2015image} limits the user's scope of action, trading their control over the segmentation outcome for more guidance by the system.
(2) Setting parameter values directly demands an insight of the user into the algorithm. 
(3)\label{sec:user_input_pictorial} Pictorial input on the image $\mathbf{I}\in\mathbb{R}^{{D_1},\,{D_2},\,{D_3}}$, 
is the most intuitive case for the user.
\mbox{$N={D_1}\cdot{D_2}\cdot{D_3}$} is the number of elements in the image.
This method mimics human behavior during knowledge transfer via a visual medium.
For the scope of this paper, a pictorial user input is utilized.
This is the most challenging class of user simulation, but also the most intuitive interaction scheme for the human operator.

According to Nickisch et al.\ \cite{nickisch2010learning}, there are three different approaches to include this user-dependent pictorial data into the evaluation process.
Given a predefined task, several human participants interact with the system in (1) user studies or by (2) crowd sourcing in order to gather plausible hints at every step of the iterative segmentation.
(3) An active user model (also called robot user) 
aims at a fast and highly scalable method to simulate plausible user interactions with the segmentation system.
Such a model may be learned from a sufficiently large user interaction database compiled utilizing data from (1) or (2).
Alternatively, the model can be defined by a rule-based system such as \cite{nickisch2010learning,zhao2011benchmark} or the one proposed here. 

\subsection{State-of-the-art}

Upconvolutional network topologies such as the \mbox{FCN} are a promising technique for solving element-wise (dense) prediction problems on image data \cite{long2015fully,lecun2015deep,wurfl2016deep}. 
Classical convolutional neural networks \mbox{(CNNs)} typically append fully-connected layers or multilayer perceptrons to their contracting path.
In contrast, \mbox{FCNs} solely consist of convolutional and pooling layers.
The missing layer types are substituted by unpooling/upsampling and deconvolutional/upconvolutional layers in an expanding path.
Shift-invariant filter operations are therefore applied in each step of the segmentation computation, forming hierarchies of learned features.
In this paper, we utilize the \mbox{FCN} topology for pixel level classification, that is commonly referred to as the {U-net} \cite{ronneberger2015u}.
In \mbox{CNNs}, pooling is performed to introduce a hierarchy of features, preserving only a condensed version of the former neighborhood's information.
Some localization information is lost during each pooling operation due to an increasingly coarser image representation.
The \mbox{U-net} architecture recovers spatial information 
by preserving spatial resolution from previous layers and linking it to later neurons in less fine-grained layers.
The \mbox{U-net} architecture in combination with augmentation of the input image data \cite{simard2003best} allows for particularly high accuracy segmentation results from a relatively small set of training data.

\mbox{FCNs} have been successfully applied to several segmentation tasks \cite{long2015fully,ronneberger2015u}, but were so far only considered in a fully automated context, consequently, omitting valuable prior knowledge of trained personnel.
In this paper, we extend the use of \mbox{FCNs} by an interactive component during training.
The resulting fully trained network is then able to improve its segmentation suggestions depending on user-defined seed points during the segmentation process.
This property is achieved by simulating plausible user inputs during the training phase of the artificial neural network (\mbox{ANN}) by an active user model which reacts to segmentation suggestions.
The interactive learning-based system is evaluated w.\,r.\,t.\ the well-known \mbox{U-net} \cite{ronneberger2015u} without a user model and \mbox{GrowCut} \cite{vezhnevets2005growcut,amrehn2016comparative} segmentation methods via the S{\o}rensen-Dice coefficient (Dice) \cite{dice1945measures}.

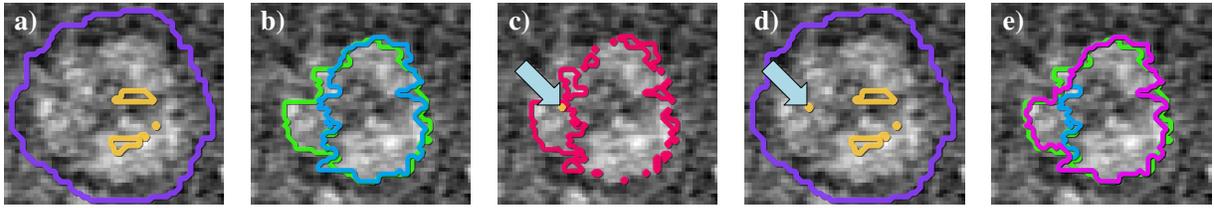
\begin{figure*}
	\centering
	\resizebox{0.942\textwidth}{!}{%
		\Large
		\begin{tabular}{lllll}
			\begin{tikzpicture}
			\node[anchor=south west,inner sep=0] (image) at (0,0) {%
				\includegraphics[trim={0 0 0 35},clip,height=0.1722\textwidth]{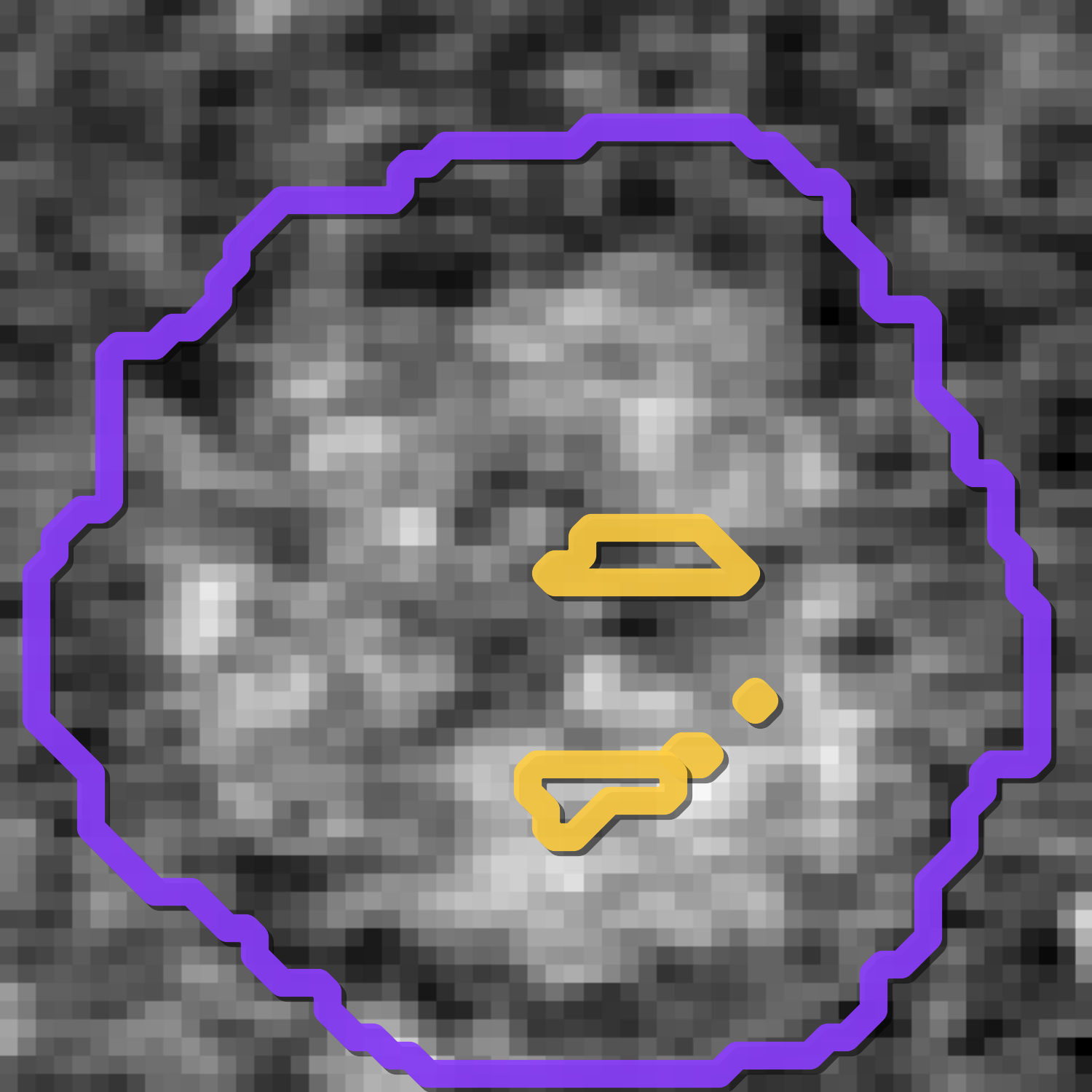}%
			};
			\begin{scope}[x={(image.south east)},y={(image.north west)}]
			\node[white,font=\bfseries] at (0.1,0.9) {a)};
			\end{scope}
			\end{tikzpicture} &
			\begin{tikzpicture}
			\node[anchor=south west,inner sep=0] (image) at (0,0) {%
				\includegraphics[trim={0 0 0 35},clip,height=0.1722\textwidth]{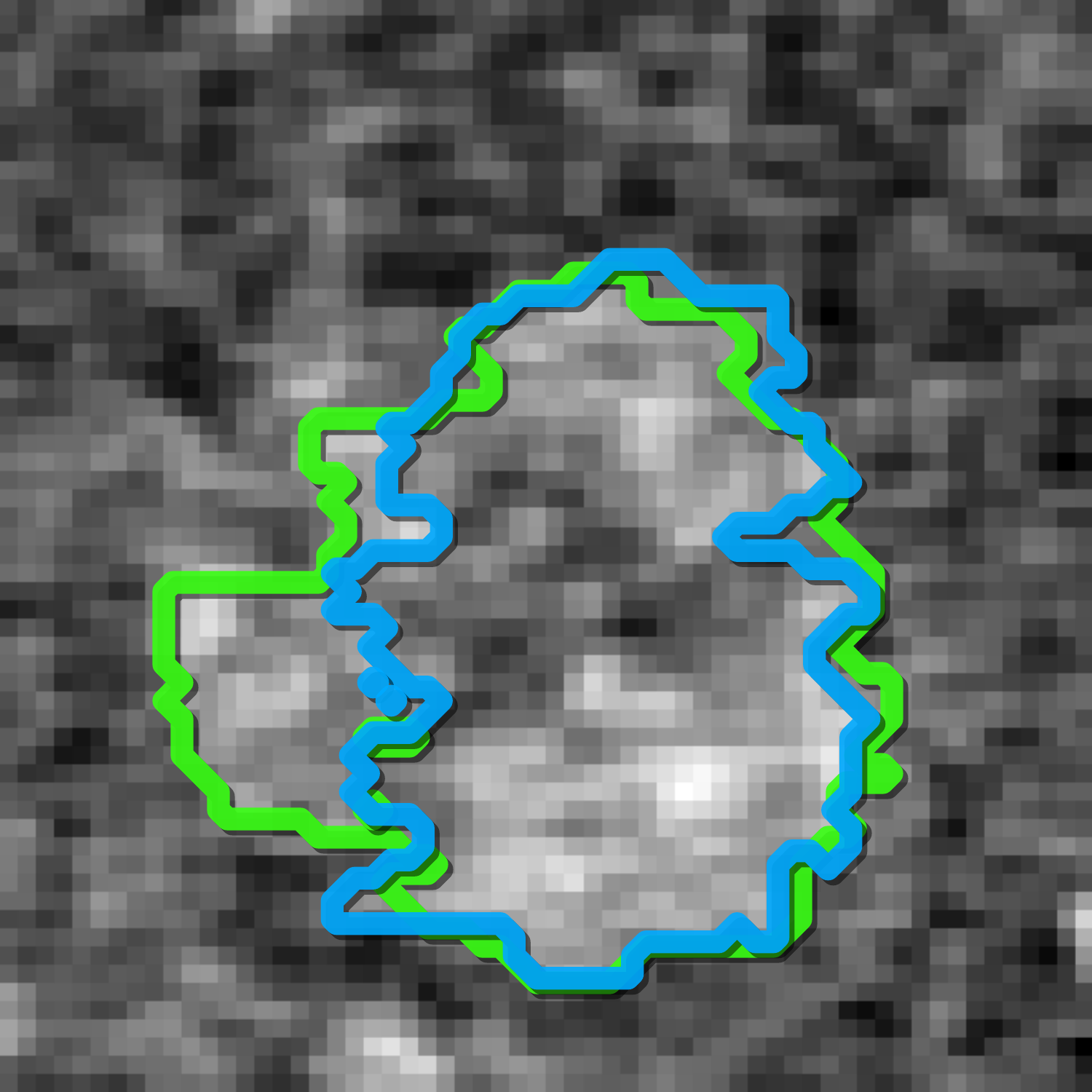}%
			};
			\begin{scope}[x={(image.south east)},y={(image.north west)}]
			\node[white,font=\bfseries] at (0.1,0.9) {b)};
			\end{scope}
			\end{tikzpicture} &
			\begin{tikzpicture}
			\node[anchor=south west,inner sep=0] (image) at (0,0) {%
				\includegraphics[trim={0 0 0 35},clip,height=0.1722\textwidth]{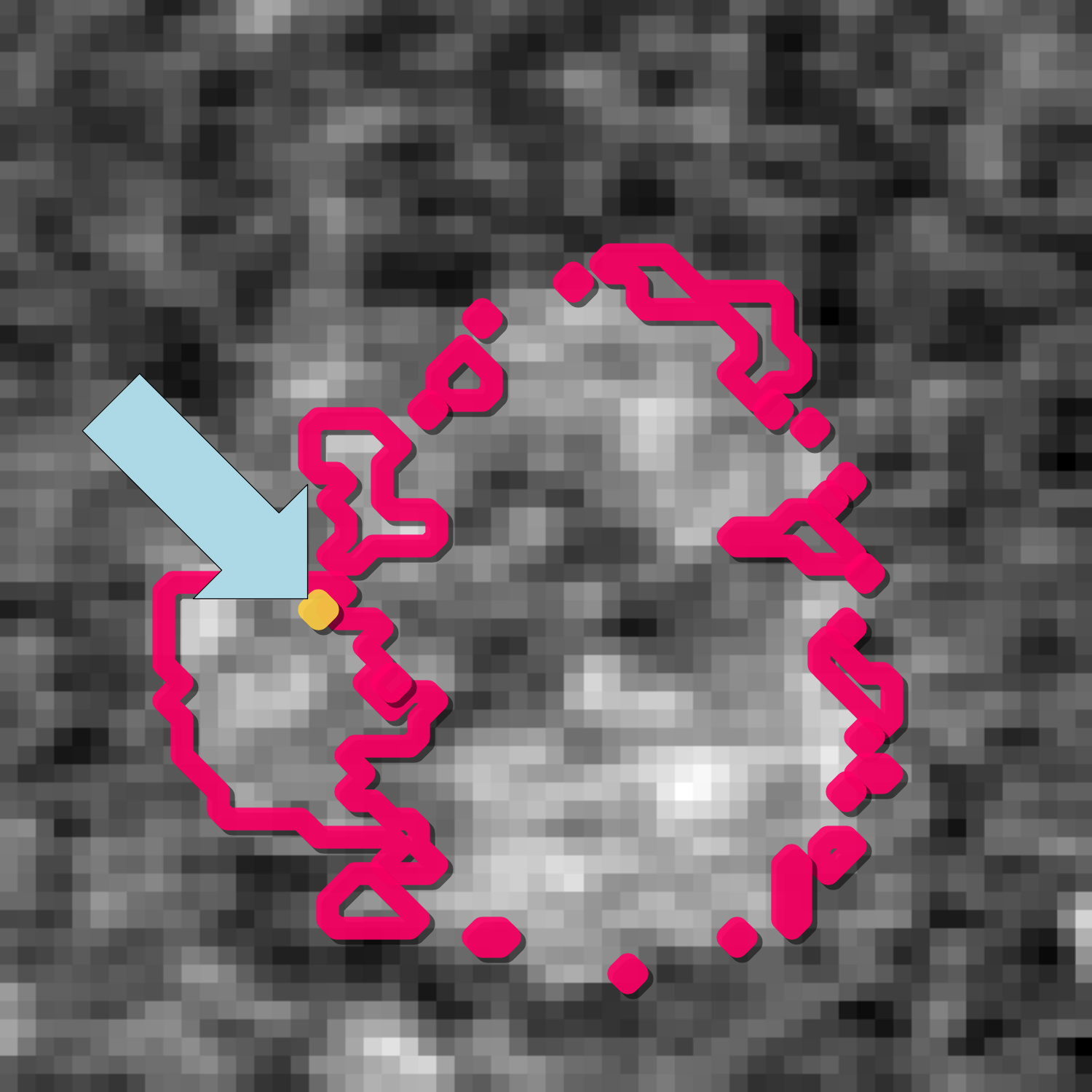}%
			};
			\begin{scope}[x={(image.south east)},y={(image.north west)}]
			\node[white,font=\bfseries] at (0.1,0.9) {c)};
			\end{scope}
			\end{tikzpicture} &
			\begin{tikzpicture}
			\node[anchor=south west,inner sep=0] (image) at (0,0) {%
				\includegraphics[trim={0 0 0 35},clip,height=0.1722\textwidth]{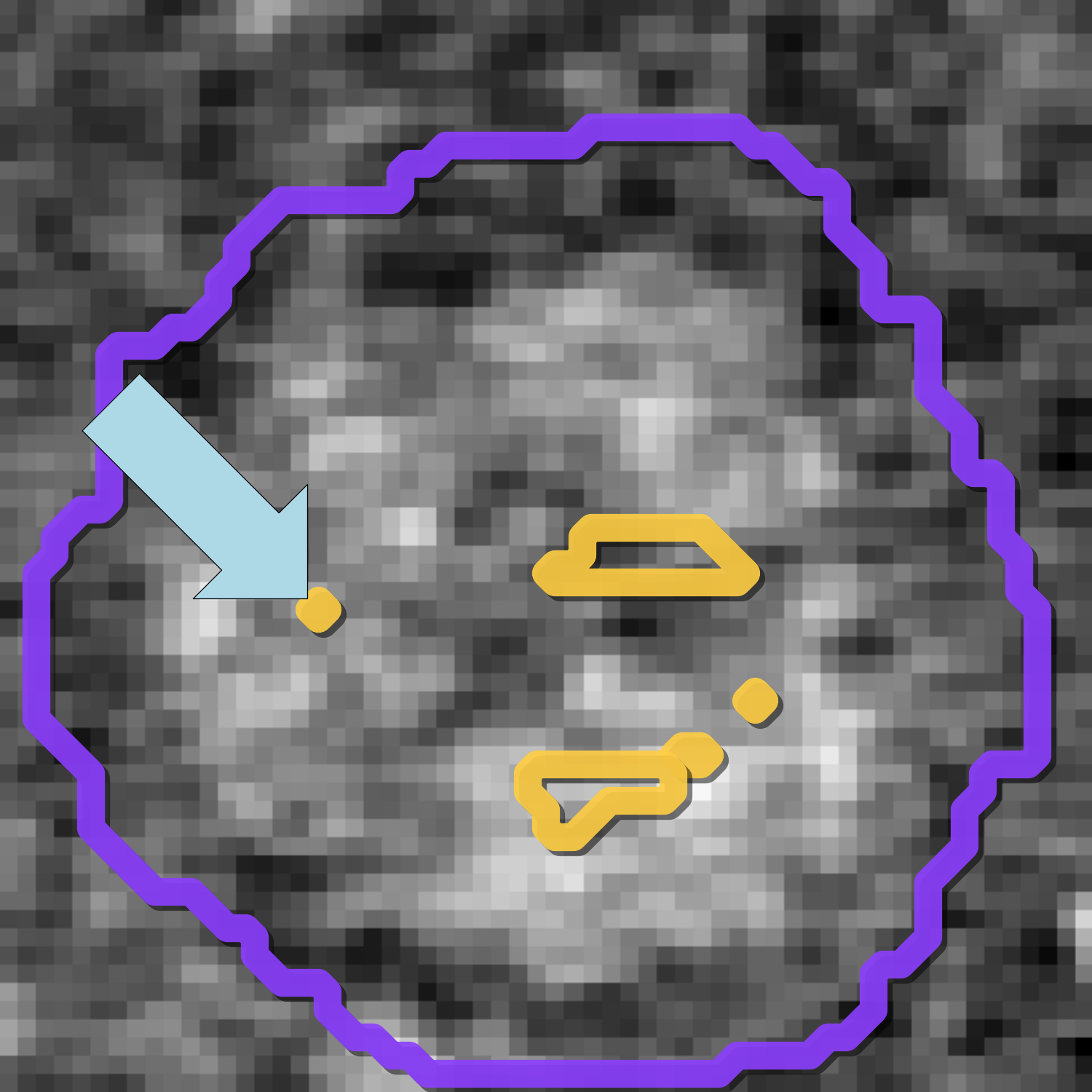}%
			};
			\begin{scope}[x={(image.south east)},y={(image.north west)}]
			\node[white,font=\bfseries] at (0.1,0.9) {d)};
			\end{scope}
			\end{tikzpicture} &
			\begin{tikzpicture}
			\node[anchor=south west,inner sep=0] (image) at (0,0) {%
				\includegraphics[trim={0 0 0 35},clip,height=0.1722\textwidth]{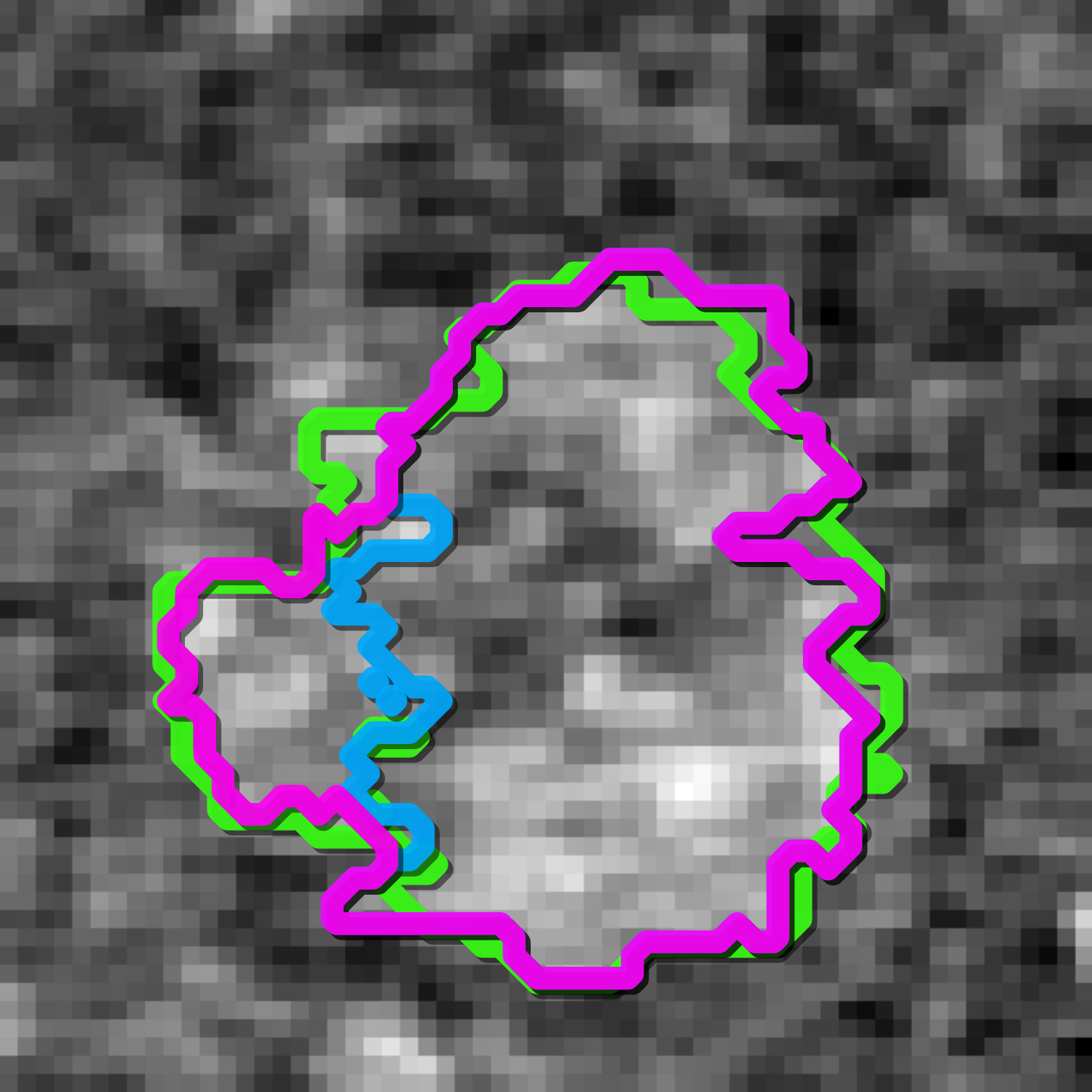}%
			};
			\begin{scope}[x={(image.south east)},y={(image.north west)}]
			\node[white,font=\bfseries] at (0.1,0.9) {e)};
			\end{scope}
			\end{tikzpicture}
		\end{tabular}%
	}%
	\caption{Active user model: (a) from the current seed mask $\mathbf{M}_{\mathbf{S}^{0}_b}$, (b) a segmentation $\mathbf{M}_{\mathbf{B}^{0}}$ is computed (cyan). Ground truth $\mathbf{M}_\mathbf{G}$ is depicted in green. (c) The difference mask $\mathbf{M}_{\mathbf{E}^{0}}$ (red) is used to randomly select misclassified image elements, in this case a single element $\mathbf{s}_i$. (d) The seed mask is updated by the user model. (e) Improved segmentation $\mathbf{M}_{\mathbf{B}^{1}}$ is obtained (magenta) w.\,r.\,t.\ the previous segmentation (b) in blue.
	}%
	\label{fig:robot_seeding_example}%
\end{figure*}

\section{Methods}

\subsection{Interactive Network Architecture}

Pictorial scribbles (seed points, lines, and shapes) are drawn by the user as an overlay mask $\mathbf{M}$ on the visualization of the image $\mathbf{I}$ to segment.
Lines and complex shapes are represented as a set of seed points.
A seed point denotes a tuple $\mathbf{s}_i:=(p_i,\,l_i)$ where $p_i\in[0,\,N)$ is a position in the image space and $l_i\in\{$background$,\,$foreground$\}$ represents the label at this position in a binary segmentation system.
Seed points are defined by the user in order to act as a representative subset $\mathbf{S}$ of the segmentation ground truth $\mathbf{G}:=\{\mathbf{s}_0, \,\dots,\,\mathbf{s}_{N-1}\}$, $\mathbf{S}\subseteq\mathbf{G}$.
The image with same dimensions as $\mathbf{I}$ and values $l_i$ at image coordinates $p_i$, where tuple $(p_i,\,l_i)\in\mathbf{S}$, is called the seed mask $\mathbf{M}_\mathbf{S}$. 
In each iteration, active user models add labeled scribbles to $\mathbf{S}$ based on the difference of the current segmentation to the ground truth in order to define the next interaction with the system, a strategy human users pursue as well.

\subsection{User Model}\label{sec:user_model}

We propose a rule-based user model as a surrogate operator of the interactive segmentation system during training.
The user model simulates a human user during the training phase of the neural network by altering the input of the network for each epoch $t$.
User input is considered additive.

For the initial interaction $\mathbf{S}^{t=0}_b:=\mathbf{M}_{\mathbf{G}_e^b}\cup\mathbf{M}_{\mathbf{G}_d^{b}}$, binary erosion 
and binary dilation
are performed on the voxel data $\mathbf{M}_\mathbf{G}$ \cite{rhemann2009perceptually}.
The foreground labeled image elements after $b$ iterations of foreground erosion $\mathbf{M}_{\mathbf{G}_e^b}$, 
are combined with the background labeled image elements after $b$ iterations of foreground dilation $\mathbf{M}_\mathbf{G_d^b}$.
This method prevents initial seed placement near the true contour line and mimics a quickly drawn rough estimate of the object to segment, 
as shown in \mbox{Fig.\,\ref{fig:robot_seeding_example}\,(a)}, where the outline of $\mathbf{S}^{t=0}_{b=5}$ is depicted.

\begin{figure*}
	\centering
	\includegraphics[width=0.77\textwidth]{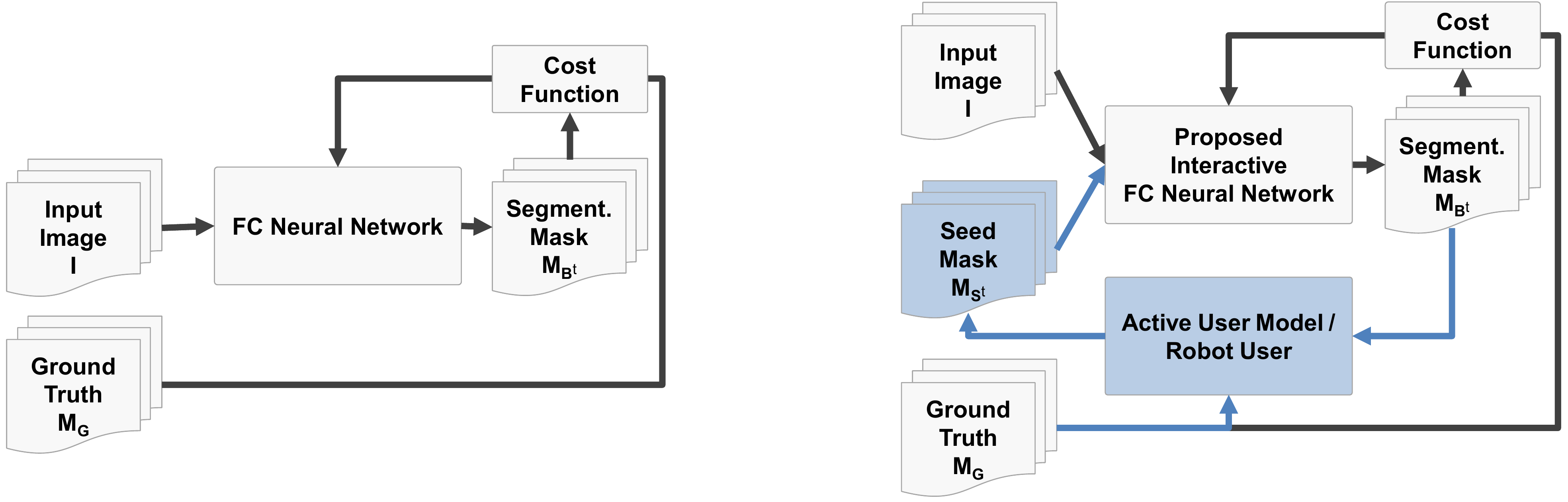} %
	\caption{Traditional FCN training procedure (left) and proposed training method by user simulation (right).}%
	\label{fig:seeding_schematically}%
\end{figure*}

\begin{figure*}%
	\centering
	\includegraphics[width=0.77\textwidth]{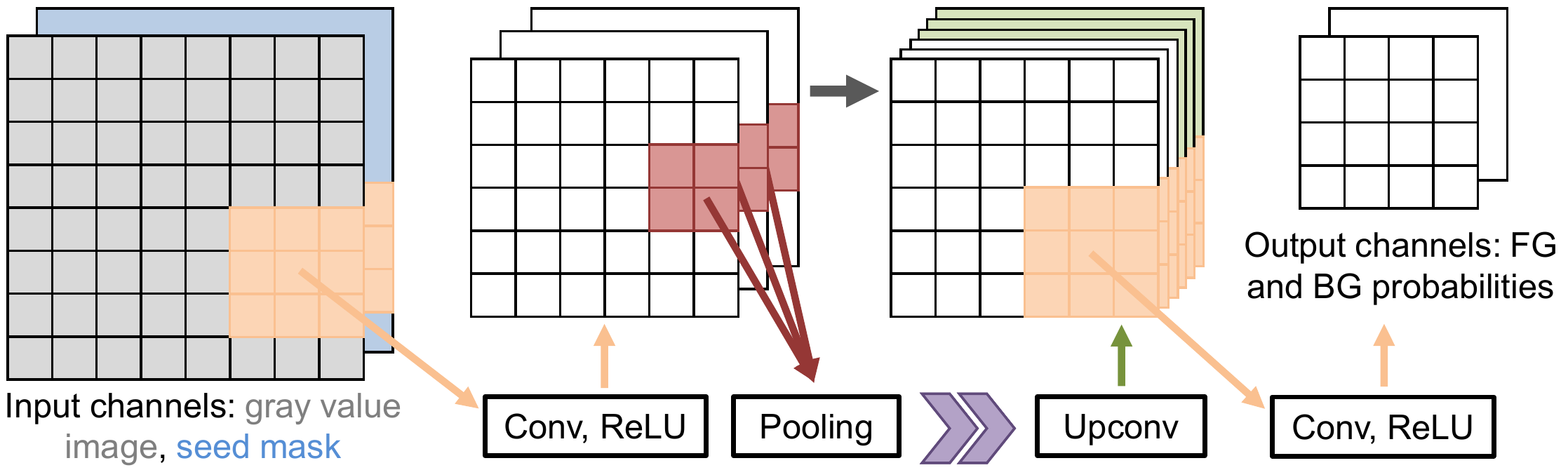}%
	\caption{Schematic \mbox{FCN} computation
		including user information as additional input (blue). Purple arrows represent further computational layers based on \cite{ronneberger2015u} topology.}%
	\label{fig:ui-net_topology}%
\end{figure*}

At iterations $t>0$, the active user model takes the current binary segmentation 
\mbox{$\mathbf{B}^t:=\{\mathbf{s}^t_0, \,\dots,\,\mathbf{s}^t_{N-1}\}$} 
and ground truth $\mathbf{G}$ as input, as depicted in \mbox{Fig.\,\ref{fig:robot_seeding_example}\,(b)}. 
The user model extracts the set \mbox{$\mathbf{E}^t:=\displaystyle\mathbf{G}\setminus (\mathbf{B}^t\cup\mathbf{S}^t)=\{\mathbf{G}\ni\mathbf{s}_i\notin(\mathbf{B}^t\cup\mathbf{S}^t)\}$} of incorrect label assignments.
It selects a subset 
\mbox{$\mathbf{\emptyset}\neq\mathbf{U}^t_n\subseteq\mathbf{E}^t;$} \mbox{$|\mathbf{U}^t_n|=\lceil|\mathbf{E}^t|\cdot n\rceil$}, where \mbox{$0<n<1$}, 
uniformly at random (\mbox{Fig.\,\ref{fig:robot_seeding_example}\,(c)}).
Subsequently, these seed points are added to the current seeds $\mathbf{S}^t$ to create \mbox{$\mathbf{S}^{t+1}:=\mathbf{S}^t\cup\mathbf{U}^t_n$}.
\mbox{Fig.\,\ref{fig:robot_seeding_example}\,(d)} depicts a choice of seeds $\mathbf{S}^1$, which is utilized to generate an improved segmentation $\mathbf{B}^1$ in \mbox{Fig.\,\ref{fig:robot_seeding_example}\,(e)}.
As shown in \mbox{Fig.\,\ref{fig:seeding_schematically}\,(right)}, the proposed neural network uses the gray-valued image data as input as well as user information in form of the seed mask $\mathbf{M}_{\mathbf{S}^t}$ for each input image.
$\mathbf{I}$ and $\mathbf{M}_{\mathbf{S}^t}$ are incorporated into the system as two separate input channels as depicted in \mbox{Fig.\,\ref{fig:ui-net_topology}}.

\section{Experiments} 

\subsection{Data}

The data set used in this paper consists of $27$ volumetric images.
They correspond to reconstructions of \mbox{DYNA-CT} acquisitions of human patients' abdomina, with voxel resolutions from \mbox{$0.46^3\,{mm}^3$} to \mbox{$0.68^3\,{mm}^3$}.
The hepatic lesions are fully annotated 
by two medical experts 
and can be fully embedded in manually selected cubic volumes of \mbox{$100^3$} voxels.
This defines the fixed output size of the \mbox{FCN}.
For the input volumes of interest (\mbox{VOI}), the dimensions of the output images have to be increased, 
to compensate the reduction of input image dimensions in each consecutive hidden layer, 
due to the border handling during individual convolution operations.
Since the lesions are not on the border of the abdominal image volume, these cubic volumes can be padded with $92$ voxels of surrounding gray-valued image data to \mbox{VOIs} of $284^3$ voxels. 
Due to the small amount of fully annotated data sets available and to reduce the time of the learning process as well as the number of trainable weights of the system, we use 2-D slices of the 3-D volumes as input for the \mbox{FCN}. Therefore, parts of the spatial context information is neglected by the current system.
The \mbox{VOI} cubes are sliced in transverse, coronal, and sagittal orientations.
Planes which do not contain any tumor object information are discarded to preserve a more balanced label distribution over all input slices.
The data is divided (per patient and volumetric image) into $6,375$ 
2-D images for training, $1,875$ 
for validation, and $1,800$ images for testing ($10,050$ in total).

\subsection{\mbox{UI-Net} Parameters}

\noindent %
We decided on a network depth of $4$, $32$ initial filters of size $3\times3$, 
a batch size of $10$, and $30$ epochs for training without early-stopping.
The learning rate of $10^{-4}$ and momentum $0.9$ are set after training several FCNs on the same data set and varying parameters by an evaluation of their accuracy progression per epoch w.\,r.\,t.\ smooth\-ness, overall slope and position of the minimal validation loss value.
Data augmentation via elastic deformations \cite{simard2003best} is used to increase the amount of training and validation data by a factor of four as an additional regularizer, counteracting the risk of over-fitting during training.
A standard deviation $\sigma=4$ of the Gaussian in pixels and scaling factor $\alpha=68$ to control the deformations' intensity are chosen. 
The active user model's fraction of new input data to sample is set to $n=5\,\%$, a reasonable value to simulate a human user.
Users will not place seeds in all erroneously segmented areas of the image (manual segmentation), but rather sparsely add more seed points.
Several variants are tested in order to observe the networks behavior given more domain knowledge via the user input channel. 
The values for a user input mask are set to $-1$ for background, $0$ for undecided, and $1$ for foreground.
Due to the same distance to the $0$ value, the network does not inherently favor object or background labels while computing weighted sums during training.
Values of $\mathbf{I}$ are normalized between $0$ and $1$ accordingly. 

\subsection{\mbox{UI-Net} Seeds}

Three experiments are conducted utilizing the \mbox{UI-net} architecture:
(1) A varying number \mbox{$b\in [5,\,15]$} of erosion and dilation operations are used to generate inputs $\mathbf{S}^{t=0}_b$ in order to examine segmentation quality w.\,r.\,t.\ additional domain knowledge inserted into the input layer. 
The smaller $b$, the more information is provided.
During training, no update step by the user model is utilized here, since the networks are trained with the same data in each epoch.
We will refer to a network with property \mbox{$\mathbf{S}^{t=i}=\mathbf{S}^{t={i+1}}$} as \emph{static} during training. 
(2) To infer, whether additional input data provided by the rule-based user model improves the segmentation quality,
seeds are generated with an alternative system to (1).
A varying number \mbox{$|\mathbf{G}|\cdot n$} of seeds ($\mathbf{S}^{t=0}_n$) are sampled uniformly at random from \mbox{$\mathbf{G}$}, where \mbox{$n\in [0.05,\,0.9]$}.
Static training is used.
(3) Multiple simulated user interactions with 
\mbox{UI-nets} are evaluated.
The \mbox{UI-nets} are trained with $\mathbf{S}^{t=0}_{b=10}$ as initial seeds and the user model proposed in Sec.\ \ref{sec:user_model}.

\section{Results and Discussion} 
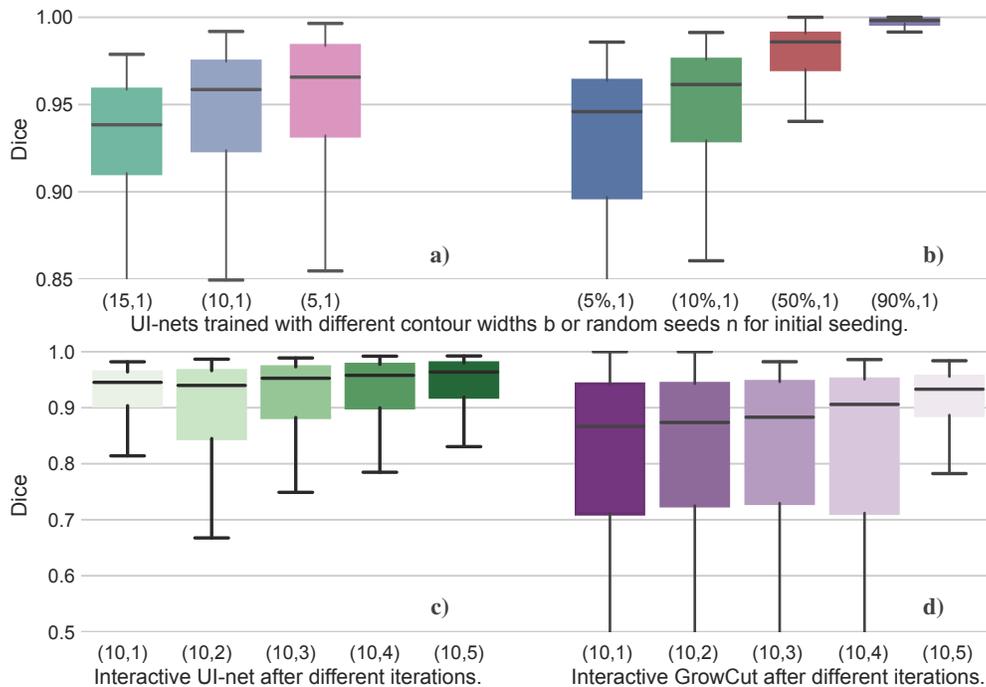
\begin{figure*}[t]
	\centering
	{
		\begin{tikzpicture}%
		\node[anchor=south west,inner sep=0] (image) at (0,0) {%
			\includegraphics[width=0.74\textwidth]{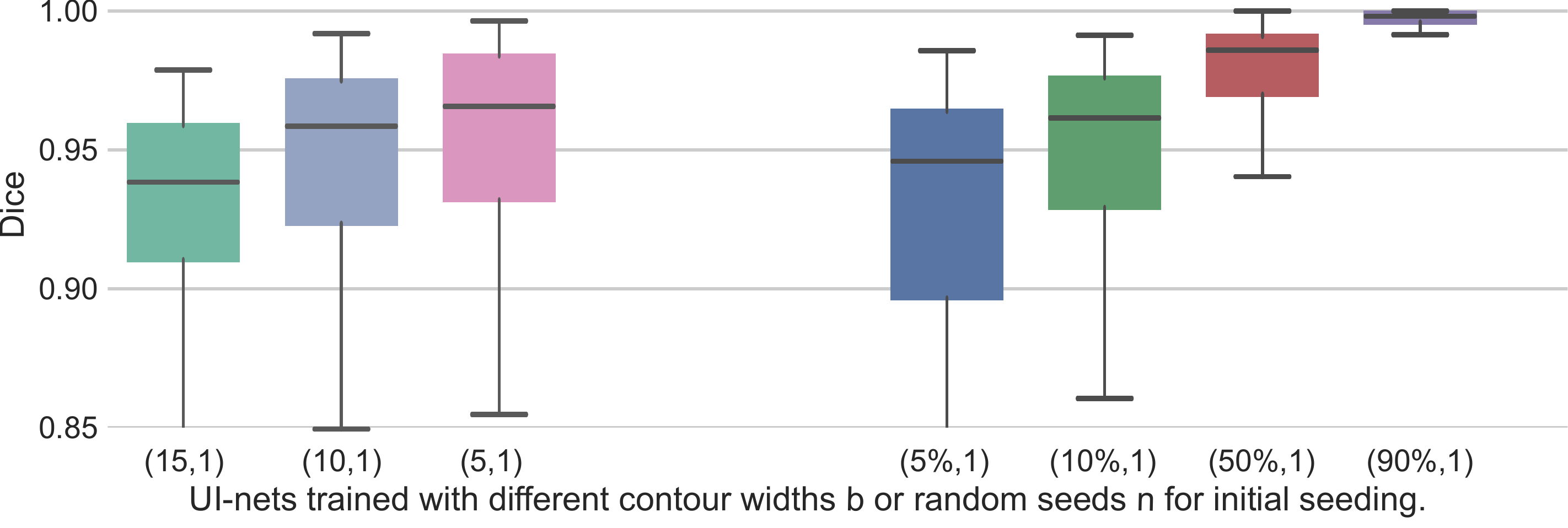}%
		};
		\begin{scope}[x={(image.south east)},y={(image.north west)}]
		\node[darkgray,font=\bfseries] at (0.435,0.259) {a)};
		\node[darkgray,font=\bfseries] at (0.935,0.259) {b)};
		\end{scope}%
		\end{tikzpicture}
		\begin{tikzpicture}%
		\node[anchor=south west,inner sep=0] (image) at (0,0) {%
			\includegraphics[width=0.74\textwidth]{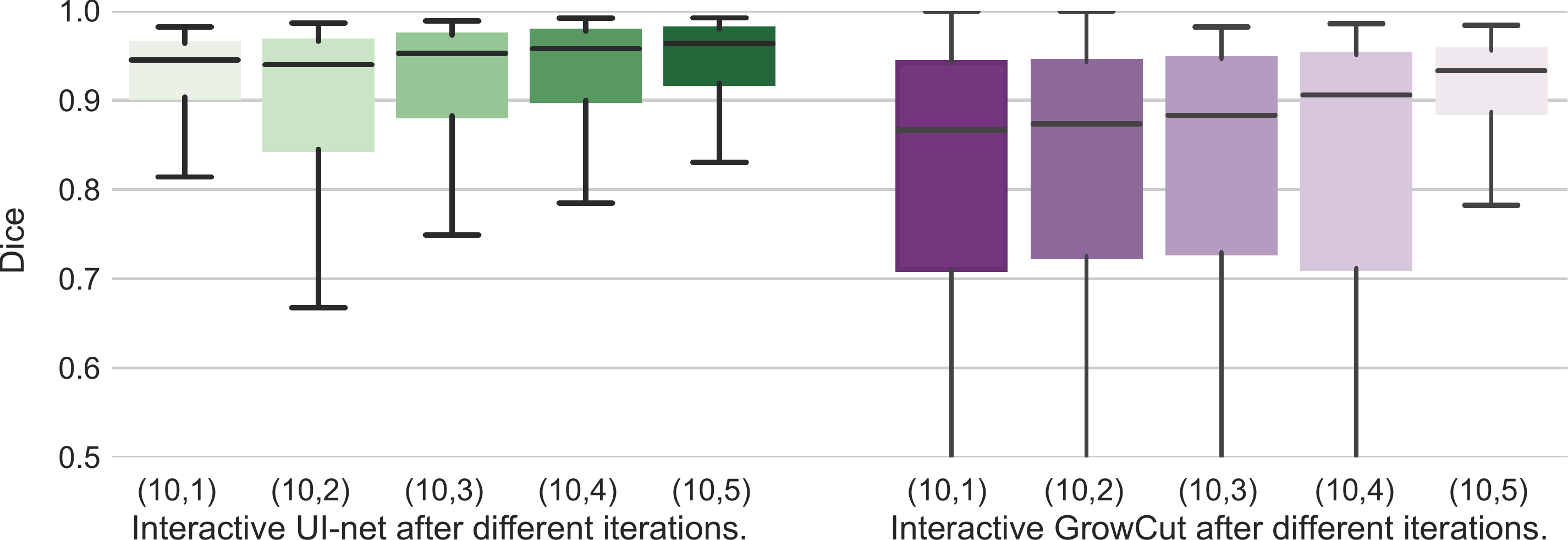}%
		};
		\begin{scope}[x={(image.south east)},y={(image.north west)}]
		\node[darkgray,font=\bfseries] at (0.435,0.225) {c)};
		\node[darkgray,font=\bfseries] at (0.935,0.225) {d)};
		\end{scope}%
		\end{tikzpicture}%
	}%
	\caption{%
		\mbox{UI-nets} trained with (a) varying contour width and (b) randomized seed masks to infer the general ability to learn from (\emph{static}) user input on first iteration data only. %
		The evaluation of several iterations, during an interactive segmentation by an active user model, is displayed in the bottom row (c,\,d). 
		The test set for iteration $1$ is always the same.
		Interactive seed changes occur only after the first iteration.
		\mbox{Legend:} \mbox{$(\langle$width$\ b\,|\,$fraction of random initial seeds$\ n\rangle$,$\langle$iteration$\rangle)$}, as in (b$=10$,\,iteration$=5$). %
	} %
	\label{fig:box_eval_user_input}
\end{figure*}

\mbox{UI-nets} were trained and evaluated with different seeding approaches.
As depicted in \mbox{Fig.\,\ref{fig:box_eval_user_input}\,(a,\,b)}, the number of given seed points correlates with the overall segmentation quality (experiments ($1$,\,$2$)).
For an evaluation with the actual user model, the interactive user input version of the \mbox{UI-net} performs best as depicted in \mbox{Fig.\,\ref{fig:box_eval_user_input}\,(c)} (experiment ($3$)).
The \mbox{UI-net} trained with an interacting user model consistently performs better with each additional input provided by the user, continuously improving its segmentation results.
Training an \mbox{FCN} with a user model which reacts on deficiencies in current segmentation results during training can therefore improve the overall segmentation result.
As visualized in \mbox{Fig.\,\ref{fig:box_eval_user_input}}\,(c,d), \mbox{UI-net} yields superior segmentation results w.\,r.\,t.\ the interactive and non-learning based \mbox{GrowCut} approach.
We found a $6\,\%$ average improvement in Dice score given the same images and active user model.

\section{Conclusion and Outlook}

We described a method to incorporate user scribbles as additional input for a semi-automatic neural network image segmentation.
A user model simulates plausible interactions of a user during the learning phase of the network.
%
In contrast to traditional \mbox{FCNs}, during each classification, new user input ground truth is generated by an operator and included into the input of the network.
The \mbox{UI-net} can be subsequently trained with this information.
The \mbox{UI-net} learns to incorporate the user information into the process of classification.
The proposed \mbox{UI-net} architecture can be superior to fully automated approaches in terms of highly accurate segmentation results, especially in medical applications where only few data sets need to be processed and only a small database of fully annotated images is available for training.
The interactive user input version would need more training epochs than a network with static user input for equivalent results, if the test setup is non-interactive as in \mbox{Fig.\,\ref{fig:box_eval_user_input}\,(a)}. %

The described technique to include user information into an \mbox{FCN} segmentation system can also be implemented via transfer learning from pre-trained non-interactive \mbox{FCNs}.
Here, a second \mbox{FCN} is trained to fine-tune the existing model with user data as an additional input besides the output of the first net.
The first \mbox{FCN} then acts as a feature extractor \cite{yosinski2014transferable} and can be trained separately from the second net for user interaction. 
The proposed user model is a rule-based system to simulate a user's behavior.
Another set of rules \cite{nickisch2010learning,zhao2011benchmark} and learning-based systems are to be evaluated for different active user models to further improve the segmentation process by reducing the amount of interactions needed from the user to achieve the same segmentation results by even more adapted 
user models utilized during training.

Disclaimer: The concept and software presented in this paper are based on research 
and are not commercially available. Due to regulatory reasons its future availability cannot be guaranteed.


\bibliographystyle{eg-alpha-doi}

\newcommand{\etalchar}[1]{$^{#1}$}


\end{document}